\pdfoutput=1

\documentclass[11pt]{article}

\usepackage[]{ACL2023}

\usepackage{times}
\usepackage{latexsym}
\usepackage{graphicx}

\usepackage[T1]{fontenc}

\usepackage[utf8]{inputenc}

\usepackage{microtype}

\usepackage{inconsolata}

\usepackage{algorithm}
\usepackage{algorithmic}
\usepackage{booktabs}
\usepackage{makecell}
\usepackage{bm}
\usepackage{hyperref}
\usepackage{amsmath}
\usepackage{multirow} 
\usepackage{color}
\DeclareMathOperator*{\argmax}{argmax}

\definecolor{lightgreen}{rgb}{0.55, 0.71, 0.0}
\definecolor{bisque}{rgb}{0.87, 0.72, 0.53}
\definecolor{lightyellow}{rgb}{0.99, 0.76, 0.0}
\definecolor{lightblue}{rgb}{0.36, 0.54, 0.66}
\definecolor{darkgray}{rgb}{0.66, 0.66, 0.66}

\usepackage{newfloat}
\usepackage{listings}

\usepackage[switch]{lineno}

\title{MemoChat: Tuning LLMs to Use Memos for Consistent Long-Range Open-Domain Conversation}

\author{\makecell{Junru Lu$^1$, Siyu An$^2$, Mingbao Lin$^2$, Gabriele Pergola$^1$, Yulan He$^3$, \\Di Yin$^2$, Xing Sun$^2$ and Yunsheng Wu$^2$} \\
  $^1$University of Warwick, $^2$Tecent YouTu Lab, $^3$King's College London\\
    $^1$\texttt{\{junru.lu, gabriele.pergola\}@warwick.ac.uk}, $^3$\texttt{yulan.he@kcl.ac.uk} \\
  $^2$\texttt{\{siyuan, marlin, endymecyyin, winfredsun, simonwu\}@tencent.com}}

\begin{document}
\maketitle
\begin{abstract}
We propose \texttt{MemoChat}, a pipeline for refining instructions that enables large language models (LLMs) to effectively employ self-composed memos for maintaining consistent long-range open-domain conversations. We demonstrate a long-range open-domain conversation through iterative ``\emph{memorization-retrieval-response}'' cycles. This requires us to carefully design tailored tuning instructions for each distinct stage. The instructions are reconstructed from a collection of public datasets to teach the LLMs to memorize and retrieve past dialogues with structured memos, leading to enhanced consistency when participating in future conversations. We invite experts to manually annotate a test set designed to evaluate the consistency of long-range conversations questions. Experiments on three testing scenarios involving both open-source and API-accessible chatbots at scale verify the efficacy of \texttt{MemoChat}, which outperforms strong baselines\footnote{Codes: \url{https://github.com/LuJunru/MemoChat}.}.
\end{abstract}


\section{Introduction}
Large language models (LLMs) have brought about a substantial revolution, fundamentally changing our lifestyle. They have emerged as a new platform that connects the realms of academia and industry within the field of artificial intelligence~\cite{zhao2023survey,yang2023harnessing}. In particular, LLMs have demonstrated an enhanced ability to synchronize effectively with human cognitive processes~\cite{du2021glm,taori2023stanford,zheng2023judging,liu2023webglm,ouyang2022training} and consequently have served as a foundation towards creating human-like conversational dialogues.

\begin{figure}[!t]
  \centering
  \includegraphics[width=\linewidth]{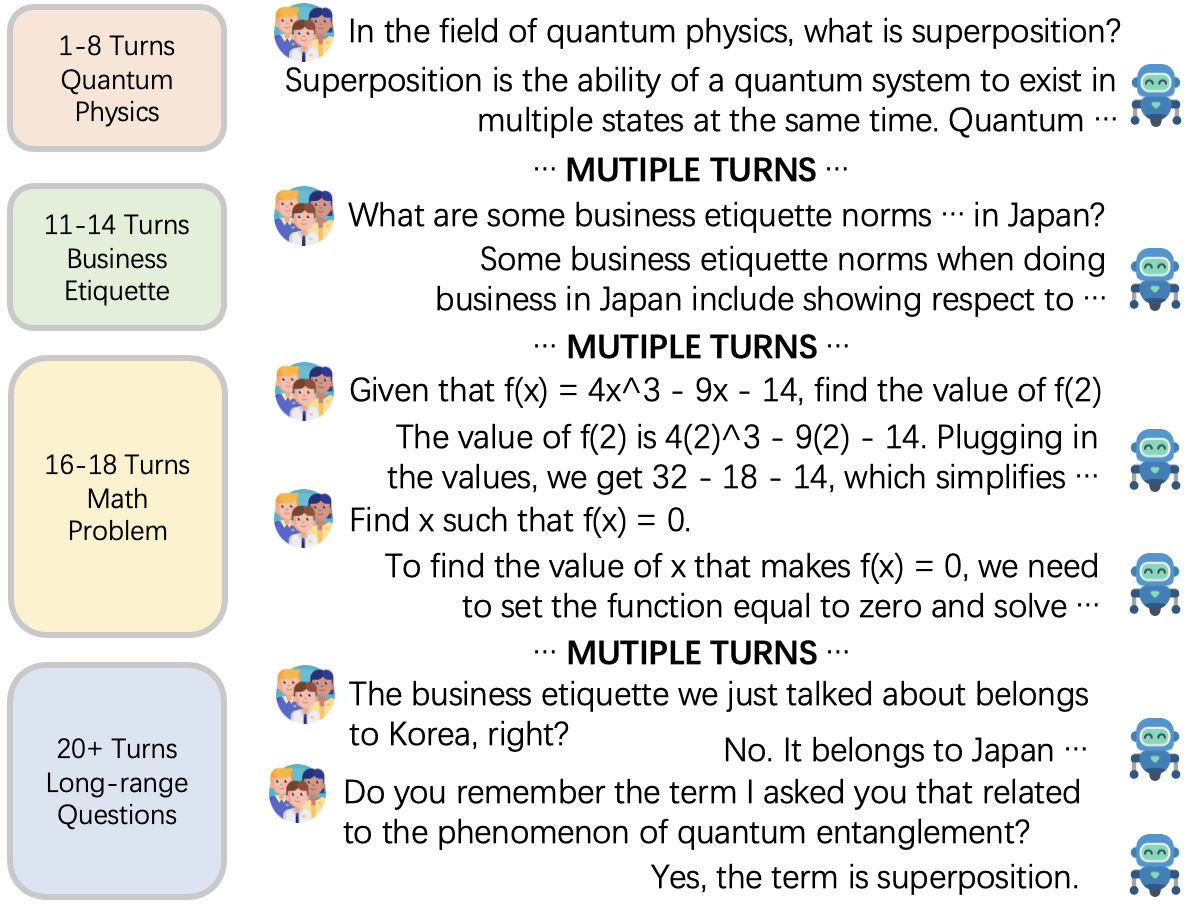}
  \caption{An example of a long-range open-domain conversation. It contains multiple topics.}
  \label{fig:sample}
\end{figure}

In contrast to the traditional dialogue systems, which mainly involve short conversations on similar topics~\cite{dinan2019wizard,komeili-etal-2022-internet,dinan2019second}, the  emerging trend in human-chatbot interaction is characterized by longer conversations encompassing diverse topics~\cite{kim2023aligning}.
Figure\,\ref{fig:sample} illustrates an example of such long-range conversations with more than 20 exchanges between a human user and a chat bot.
It involves several sequential topics such as ``\emph{Quantum Physics}'' (1 to 8 turns), ``\emph{Business Etiquette in Japan}'' (11 to 14 turns) and ``\emph{Polynomial Math Problem}'' (16 to 18 turns), while other topics are simply referred to as ``\emph{MULTIPLE TURNS}''.
Also, two topics-related long-range questions are raised subsequently, including one for evoking memories associated with ``\emph{Business Etiquette in Japan}'', and the other for retrieving historical information pertaining to ``\emph{Quantum Physics}''.
Such long-range open-domain conversations that involve diverse topics presents a challenge for conventional methods, as they struggle to effectively address the issue of retaining contextual coherence over long stretches of discourse. 
One possible strategy is to expand the input text window~\cite{tworkowski2023focused,longchat2023,ding2023longnet}. For example, positional interpolation~\cite{chen2023extending} was presented to empower RoPE-based~\cite{su2022roformer} LLMs with a text window that is 600 times longer than 
its original pretraining version. Unfortunately, the window length can not be arbitrarily expanded~\cite{beltagy2020longformer,kitaev2020reformer}.
Another potential solution is to resort to additional memories for tracking the history of conversations~\cite{zhong2023memorybank,zhao2023unimc}. For example, MPC was proposed~\cite{lee-etal-2023-prompted} to regularly store persona summaries of recent conversations as history, then use new user query $\bm{Q}$ to retrieve the top-K history $\bm{D}$ via an external DPR retriever~\cite{karpukhin2020dense}, and finally generate a new response based on both $\bm{Q}$ and $\bm{D}$. 
Similarly, an event and user portrait memory managed with FAISS~\cite{johnson2019billion} were adopted by~\cite{zhong2023memorybank}. 
Despite the advantage in retaining 
infinite contextual inputs, compared with text window extension, memory mechanism may cause accumulation errors during the retrieval of past evidence~\cite{xu2022long}.
As a result, how to enhance the accuracy of memory recall and improve the response consistency of LLM-powered chatbots in long-range open-domain conversation remains unsolved~\cite{xu2021goldfish}.

In this paper, we are particularly interested in building memory-enhanced chatbots. Instead of replying on external tools for memory creation and recall~\cite{lee-etal-2023-prompted,zhong2023memorybank,hu2023chatdb,kynoch2023recallm}, we propose a streamlined approach where we eliminate the need for these complex associate modules. Instead, we create a simplified pipeline that solely utilizes LLMs to power chatbots. 
The proposed memo-equipped pipeline, 
\texttt{MemoChat}, aims for guiding LLMs to use simple on-the-fly memos for maintaining consistent long-range open-domain conversation.
We base our approach on the concept of ``\emph{memorization-retrieval-response}'' loops within the context of long-range open-domain conversations. 
%
To achieve this, except for zero-shot learning with chatbots accessed via public APIs~\cite{openai2023gpt4,anil2023palm}, 
we reconstruct a series of instructions from publicly available datasets~\cite{adlakha-etal-2022-topiocqa,chen-etal-2021-dialogsum,ghosal2023flacuna} to fine-tune open-source LLM-powered chatbots. This fine-tuning process helps the chatbots familiarize themselves with each stage within the loop, and employ structured memos for maintaining a consistent long-range conversation. 
For fair evaluation, we invite experts to manually annotate a testing set specifically designed to evaluate the consistency of long-range conversations 
across three different types of testing scenarios. 
We verify the effectiveness of the proposed \texttt{MemoChat} through a comprehensive evaluation involving 
both a chatbot accessible via public API and four different open-source LLMs. This evaluation is carried out using a 
very strong LLM judge~\cite{zheng2023judging}.
The contributions of this paper are listed as follows:
\begin{itemize}
    \item We introduce \texttt{MemoChat}, an instruction tuning pipeline, to assist LLM-powered chatbots carrying on consistent long-range open-domain conversations through an on-the-fly memo. 
    \item We leverage several publicly available datasets to reconstruct instructions 
    to help open-source LLMs familiarize themselves with the \texttt{MemoChat} pipeline.
    \item We curate an expert-annotated evaluation set, which serves as a benchmark for assessing 
    response consistency. Experiments on both open-source and API-accessible chatbots 
    verify the efficacy of \texttt{MemoChat}.
\end{itemize}

\section{Related Work}
\subsection{Long-Range Open-Domain Conversation} 
Long-range open-domain conversation has been a widely explored task recently. Doc2Dial, a goal-oriented document-grounded dialogue dataset is proposed by ~\cite{feng2020doc2dial}, including around 48k long-range conversations with an average of 14 turns. MultiSessionChat is a dataset concerning one long-range conversation over multiple sessions ~\cite{xu2021goldfish}. 
While most of these traditional 
datasets focus on few-topic human-human scenarios, there has been a proliferation on newly multi-topic human-chatbot scenarios~\cite{kim2023aligning}.

\begin{figure*}[!t]
  \centering
  \includegraphics[width=1.0\linewidth]{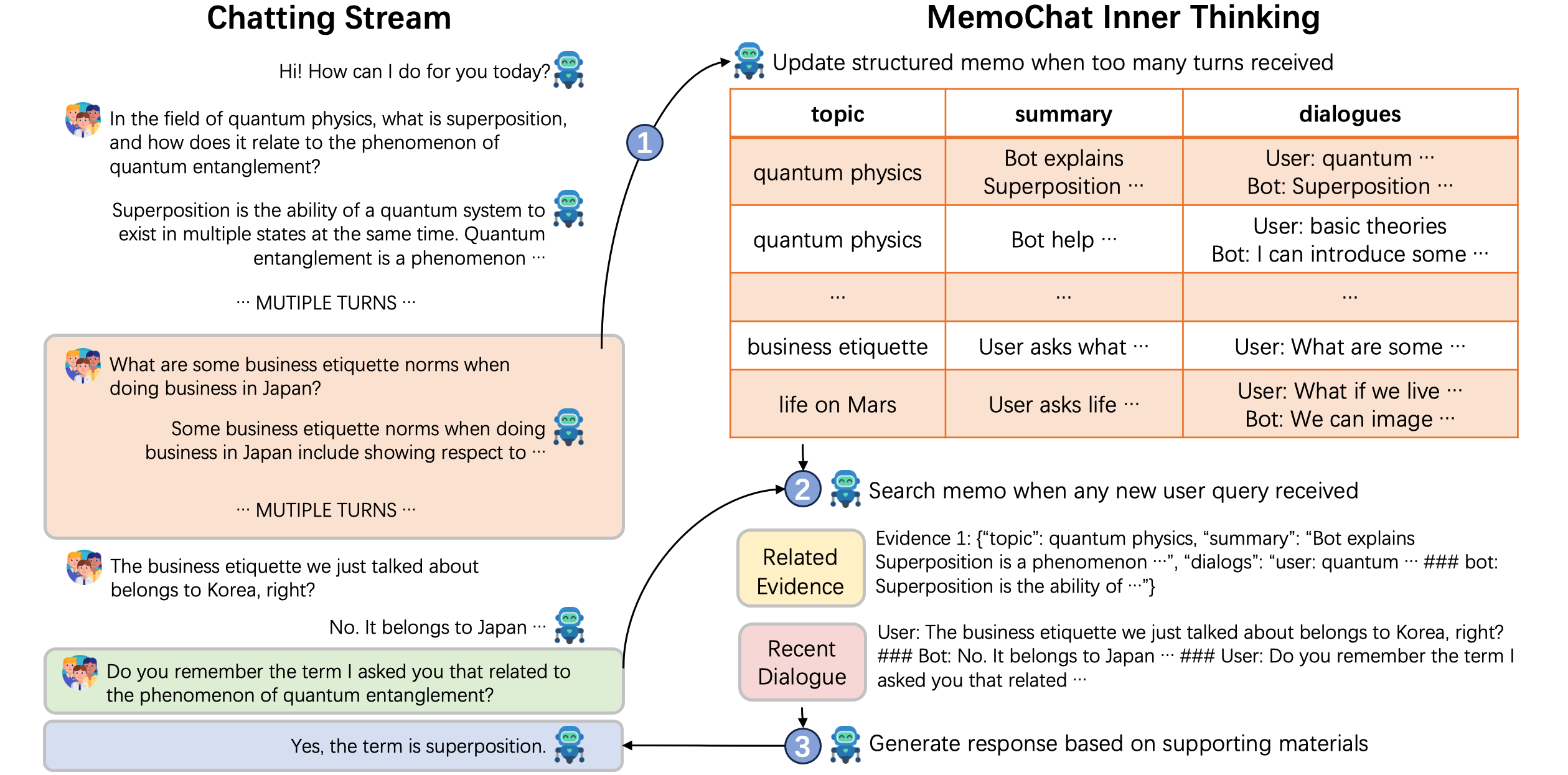}
  \caption{The overall architecture of our proposed \texttt{MemoChat} pipeline. While maintaining a chatting stream between the human user and the chatbot (left part), the chatbot will have a memo-equipped inner thinking (right part). 
  }
  \label{fig:model}
\end{figure*}

The challenges of long-range open-domain conversations lie on the answerability of open-domain questions and the capacity of long-range input understanding. External knowledge graph was incorporated to improve the answer ability of out-of-distribution open-domain questions~\cite{zhou2020kdconv}. With the development of LLMs, LLM-powered chatbots have largely improved their answering ability~\cite{shuster2022blenderbot,shuster2022language}. On the other hand, various strategies have been proposed to tackle the challenge of long-range input understanding. One possible strategy is to use additional memory to filter shorter key facts in the past long-range history to replace it. MPC~\cite{lee-etal-2023-prompted} and MemoryBank~\cite{zhong2023memorybank} were proposed to use additional memory and external retriever to manage persona, event or user portrait information. 
Our work differs from existing studies in that we solely teach the chatbot to self-use memos, and we focus on summarizing all facts in the history instead of partial persona or event facts.

\subsection{Instruction Tuning} 
Instruction tuning is a widely used strategy to fine-tune LLMs on tasks with meta-instruction ahead~\cite{wei2021finetuned}. The effectiveness of instruction tuning has been proved on many tasks, including text reasoning~\cite{sanh2021multitask}, human-bot alignment~\cite{ouyang2022training}, document ranking~\cite{qin2023large,sun2023chatgpt}, document retrieval~\cite{mueller2023metatraining}, and general conversation ability~\cite{gupta2022improving}. In this work, we spend low-cost yet carefully design instructions to teach LLM-powered chatbots for self-using on-the-fly memo to strengthen the response consistency within long-range chatting stream.


\section{Methodology}

\subsection{Formulation}

We start with formulating the long-range open-domain conversation as Generative Question Answering (GQA), in which a chatbot is expected to generate an answer $\hat{Y}=\argmax_Y p(Y|\bm{x}^{q}, \bm{x}^{h})$, given previous conversation history $\bm{x}^{h}$ and current user query $\bm{x}^{q}$.
To improve the response consistency of $\hat{Y}$, our \texttt{MemoChat} takes an step forward and proposes to decompose the long-range open-domain conversation as a three-stage ``\emph{memorization-retrieval-response}'' loop, intending to retrieve query-related evidences $\bm{x}^{h'}$ from structured memo $\bm{x}^{m}$, which is modeled as $\bm{x}^{m} = f(\bm{x}^h)$, $\bm{x}^{h'} = g(\bm{x}^q, \bm{x}^m)$, with $f(*)$ and $g(*)$ as the memo build-up and retrieval functions respectively. 
Therefore, GQA in our \texttt{MemoChat} can be reformulated as as $\hat{Y}=\argmax_Y p(Y|\bm{x}^{q}, \bm{x}^{h'})$.
%

\begin{figure*}[!t]
  \centering
  \includegraphics[width=1.0\linewidth]{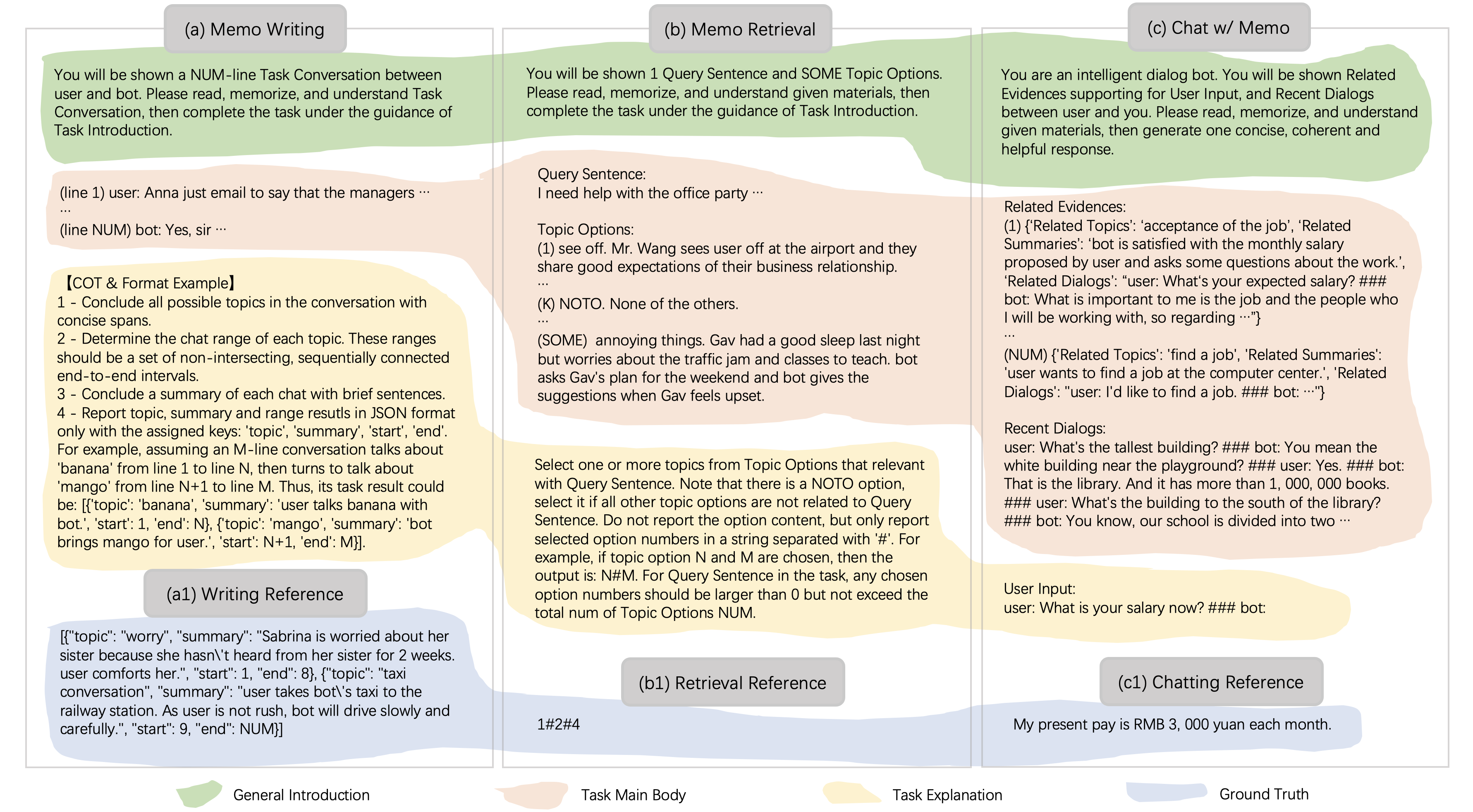}
  \caption{Instruction templates in \texttt{MemoChat} (better view in colors). Three instructions are carefully designed to serve \texttt{MemoChat} inner thinking pipeline including ``Memo Writing'', ``Memo Retrieval'' and ``Chat w/ Memo'', partitioned by {\color{darkgray}{gray}} boxes. The organization of all instructions consists of general task introduction (\emph{\color{lightgreen}{green area}}), main body of task-oriented inputs (\emph{\color{bisque}{bisque area}}), and detailed task explanation (\emph{\color{lightyellow}{yellow area}}). Ground-truth answers are provided during fine-tuning (\emph{\color{lightblue}{blue area}}). More specific examples are provided in Appendix~\ref{sec:prompt}.}
  \label{fig:prompt}
\end{figure*}

\subsection{MemoChat}
Figure\,\ref{fig:model} depicts the overall framework of our proposed \texttt{MemoChat} pipeline, which assigns the chatbot with a ``\emph{memorization-retrieval-response}'' loop of inner thinking.
In the left of Figure\,\ref{fig:model}, the user starts with the topic of ``\emph{Quantum Physics}'', then ends with ``\emph{Business Etiquette}'' topic after multiple turns of the conversation.
Very different from traditional methods that retrieve directly on these accumulated dialogues, the chatbot under our \texttt{MemoChat}, as shown on the upper-right of Figure\,\ref{fig:model}, will automatically builds and updates a structured on-the-fly memo, storing past dialogues in categories.
Then, the retrieval is conducted over all recordings according to their topics and summaries.
In the bottom-left, Figure\,\ref{fig:model} illustrates an occurrence in which the chatbot is required to link the concept of ``\emph{Quantum Physics}'' aforementioned in the very beginning of the conversation.
The chatbot manages to recall related evidences, and generate a consistent answer in bottom-right Figure\,\ref{fig:model}.

By allowing chatbots to reorganize the dialogue histories according to different topics, summarize the partial dialogue faithfully, and retrieve one or more query-related evidences quickly, \texttt{MemoChat} becomes a simple-yet-effective pipeline to guide chatbots to generate consistent responses.

\noindent\textbf{Instruction Structure}. ``\emph{Memorization-retrieval-response}'' mechanism stores the past dialogues in memo $\bm{x^m}$, recalls query-related evidences $\bm{x^{h'}}$ from memo, and generates response $\hat{Y}$ based on retrieved evidences. Different from previous literature~\cite{lee-etal-2023-prompted,zhong2023memorybank} requiring additional complex modules, we reorganize the dialogue histories into structured on-the-fly memos, such that the basic LLM-powered chatbots can simultaneously handle 
 the memo.
Thus, we develop three instructions for each stage as shown in Figure\,\ref{fig:prompt}. Overall, all instructions follow similar structure: (1) description of each stage task (\emph{\color{lightgreen}{green area}}); (2) main body of task-oriented inputs (\emph{\color{bisque}{bisque area}}); (3) explanation of each task (\emph{\color{lightyellow}{yellow area}}). The combination of three serves as the input prompts of the chatbot and ground-truth answers are provided (\emph{\color{lightblue}{blue area}}) for fine-tuning.

Herein, the ``Memo Writing'' task asks the chatbot to read a task conversation, partition the conversation based on all possible topics, write short summaries for every sub-dialogues, and write the above results to the memo in a JSON format with each instance including keys of ``topic'', ``summary'' and ``dialogues''. We consolidate the task explanation with Chain-of-Thought (CoT)~\cite{wei2023chainofthought} and In-Context Learning (ICL)~\cite{dong2023survey}. 
The ``Memo Retrieval'' task asks the chatbot to read a user query and a series of topic options, select out one or more related options and report option ID with a special format. Inspired by previous literature~\cite{rajpurkar2018know}, one ``\emph{None of the others (NOTO)}'' option is randomly inserted in the candidate options, given that the user may arbitrarily launch a new topic. To support a fast implementation, we retrieve related contents only upon the ``topics'' and ``summaries'' from the on-the-fly memo.
Finally, the ``Chat w/ Memo'' task feeds the latest query, prefixed with retrieved evidences and recent dialogues, to the chatbot. With these related evidences, the chatbot can now produce more consistent response even in the long-range open-domain conversations.

\begin{table}[t]
\resizebox{\columnwidth}{!}{%
  \begin{tabular}{l|c|c|c}
    \toprule
    \quad & \textbf{Data Size} & \textbf{Avg. Tokens} & \textbf{Sources}\\
    \midrule
    \textbf{Memo Writing} & 3,046 & 951.92 & \makecell{1,790 from T* \\ 1,256 from D*}\\
    \midrule
    \textbf{Memo Retrieval} & 3,654 & 329.86 & \makecell{1,719 from T* \\ 1,935 from D*}\\
    \midrule
    \textbf{Chat w/ Memo} & 3,300 & 273.04 & \makecell{1,602 from D* \\ 1,698 from A*}\\
    \bottomrule
  \end{tabular}}
    \caption{Statistics of our instruction training sets. ``Avg. Tokens'' stands for the average token number of each instruction. T*, D* and A* stand for TopicoQA, DialogSum and Alpaca-GPT4 dataset, respectively.}
  \label{tab:dataset}
\end{table}

\noindent\textbf{Dataset Reconstruction}. 
To implement our \texttt{Memo} pipeline, we reconstruct 10k instruction data from three public dialogue datasets including TopicoQA~\cite{adlakha-etal-2022-topiocqa}, DialogSum~\cite{chen-etal-2021-dialogsum} and Alpaca-GPT4~\cite{ghosal2023flacuna}. 
TopicoQA covers 3.9k conversations that involve 4 topics and 13 turns of dialogue on average. DialogSum covers 13k conversations that focus on 1 topic across 5.5 turns on average, and provides manual summary for each conversation. Both TopicoQA and DialogSum use human-human conversations in spoken style, and focus on daily life scenes. 
While Alpaca-GPT4 is a human-chatbot 1-turn conversation dataset created by running GPT4 API~\cite{openai2023gpt4} on the Alpaca dataset~\cite{taori2023stanford}. It covers more sophisticated topics in written style. We sample 1.6k cases to balance the instructions from another two datasets.

Table\,\ref{tab:dataset} reports the statistics and sources of the three instruction training sets. 
The ``Memo Writing'' set consists of 1,790 instances from TopicoQA and 1,256 samples from DialogSum. Given the unavailability of ``summary'' information, we only construct ``topic'' and ``dialogues'' in the memo for conversations from TopicoQA dataset.
Considering only one topic in each conversation of DialogSum, we combine a few variable set of conversations to mimic a multi-topic long-range conversation.
Similar strategies are adopted to form ``Memo Retrieval'' set that contains 1,719 and 1,935 instructions from TopicoQA and DialogSum, respectively. For each retrieval instruction, its query sentence is made up with several sequential sentences from original conversation and related topics are extracted as options. The ``\emph{NOTO}'' option is inserted with a probability of 10\%.
The ``Chat w/ Memo'' set comprises 1,602 and 1,698 instructions from DialogSum and Alpaca-GPT4. The former aims at generating instructions with related evidences, and the latter generates instructions without any related evidences to fit with unseen topics.

In conclusion, we succeed in building three training sets to achieve our ``memorization-retrieval-response'' loop. Besides, a total of 400 evaluation instructions are similarly constructed from the testing sets of the source datasets, including 160, 82 and 65 samples for ``Memo Writing'', ``Memo Retrieval'' and ``Chat w/ Memo'', respectively.


\noindent\textbf{Challenges}. To our best knowledge, we are the first to build such a group of systematic sequential instructions. 
We report major challenges for any potential future follow-ups. Some specific examples are available in Appendix~\ref{sec:challenge}.

The first challenge faceed is ``Prompt Copy'' in which the LLMs found a special shortcut to directly copy format examples in the prompts as the answer. To avoid this, we suggest to use dummy variables to replace any numerical values to avoid possible answer leaking. The second challenge refers to ``Catastrophic Forgetting'' in which the chatbots lean towards generating early-stopped or repeated answers if the volume of ``Chat w/ Memo'' is less than ``Memo Writing'' or ``Memo Retrieval''. A balanced volume is encouraged by mixing with certain amount of dialogue corpus to maintain the chatting ability~\cite{ghosal2023flacuna}. We consider Alpaca-GPT4 in our construction. The third challenge lies in ``Prompt Misplacement'' in which the chatbots 
performance degrade when swapping the input order of the main body of task-oriented inputs (\emph{\color{bisque}{bisque area}} in Figure\,\ref{fig:prompt}) and explanation of each task (\emph{\color{lightyellow}{yellow area}} in Figure\,\ref{fig:prompt}). Task explanation 
attached at the end is crucial for chatbots to understand instructions better~\cite{liu2023lost}.

\section{Experiments}
\subsection{Experimental Setting} 

\noindent\textbf{Datasets}. Except for the reconstructed instruction datasets, we also invite three experts to manually annotate a new long-range open-domain conversation dataset to validate the response consistency.
The annotated dataset, referred to as MT-Bench+, is built based on MT-Bench~\cite{zheng2023judging}, a human-chatbot testing dataset with a total of 80 two-turn open-domain questions, covering topics such as ``STEM exams'' and ``literary writing''. This is achieved by:
(1) expanding each of the two-turn questions to a four-turn version by the experts; 
(2) sampling 2$\sim$4 questions from every 3$\sim$4 of four-turn versions to form a long-range chatting stream with a total of 12$\sim$15 turns;
(3) adding long-range questions to the end of the chatting streams for ``Retrospection'', ``Continuation'' and ``Conjunction'' by the experts.
Herein, ``Retrospection'' is restricted to the aforementioned terms related questions  (\emph{e.g.}, span extraction), ``Continuation'' makes a further task based on previous knowledge (\emph{e.g.}, story completion), and ``Conjunction'' is to ask a cross-topic response (\emph{e.g.}, redo task 2 in reference to task 1). 
We build 18 questions for each of the three types in MT-Bench+, 54 in total.

\noindent\textbf{Baselines}. We consider the popular Fastchat-T5-3b and Vicuna-7b/13b/33b~\cite{zheng2023judging}, which are human-aligned versions of Flan-T5~\cite{chung2022scaling} and Llama~\cite{touvron2023llama} with massive human-bot dialogues~\cite{kim2023aligning}, respectively, as the baseline models. We further fine-tune them on our reconstructed instruction data to form our \texttt{MemoChat}.
For all experiments, we set the text window to 2k to follow off-the-shelf conventions~\cite{workshop2023bloom}. However, our \texttt{MemoChat} keeps the flexibility and scalability to larger windows as chatbots can automatically adjust memory range.
%
%
%
As for downstream consistency testing, 
one public API and two open-source memory-assistant chatbots are included as baselines:
\begin{itemize}
    \item \textbf{ChatGPT-2k}. We use the same 2k text window to arrange the input of gpt-3.5-turbo API.
    \item \textbf{MPC-ChatGPT}. Plug MPC~\cite{lee-etal-2023-prompted} pipeline with ChatGPT-2k. 
    \item \textbf{MemoryBank-ChatGPT}. Plug MemoryBank~\cite{zhong2023memorybank} with ChatGPT-2k.
\end{itemize}

GPT4 API is excluded as GPT4-generated data are included in the training set, and we rely on GPT4 for evaluation in downstream chatting test.

\noindent\textbf{Metrics}. We adopt four groups of metrics to evaluate the performance of intermediate instruction tuning and the downstream consistency checking. 
%
Since we ask chatbots to partition dialogue histories to build memo, in reference to Name Entity Recognition (NER)~\cite{ding-etal-2021-nerd}, we therefore evaluate generated topics and summaries in ``Memo Writing'' task with precision ($\bm{P}$), recall ($\bm{R}$) and micro F1-score ($\bm{F_1}$) only when predicted sub-session range and ground truth matches. In detail, the $\bm{P}$ score of topic checking uses exact match following NER, while the $\bm{P}$ score of summary checking is replaced with float BertScore~\cite{zhang2019bertscore} number varying between 0 to 1. We use regular F1-score ($\bm{F_1}$) on ``Memo Retrieval'', and use BertScore to check the semantic similarity between generated response and golden references in ``Chat w/ Memo'' task. As for downstream response consistency checking, inspired by LLM-as-a-judge strategy~\cite{zheng2023judging}, we select out related dialogue histories, and the long-range checking question with  corresponding response, then use GPT4 to score the faithfulness with an integer scale 1$\sim$100. It has been explored in~\cite{pan2023rewards} to show that credible LLMs such as GPT4 have a comparable analysis ability to human.

\begin{table}[t]
\resizebox{\columnwidth}{!}{%
  \begin{tabular}{c|c|c|c|c}
    \toprule
    \quad & \makecell{\textbf{Fastchat-}\\\textbf{T5-3B}} & \textbf{Vicuna-7B} & \textbf{Vicuna-13B} & \textbf{Vicuna-33B}\\
    \midrule
    \textbf{Batch} & 8 & 16 & 8 & 4\\
    \midrule
    \textbf{Accumulations} & 2 & 1 & 2 & 4\\
    \midrule
    \textbf{HyperParams} & \multicolumn{4}{c}{\makecell{Epoch=3, Global Batch=128, Seq Len=2048, \\Lr=2e-5, Warmup Ratio=0.04, Gen Temperature=0.2}}\\
    \midrule
    \textbf{CPU RAM (GB)} & 73.01 & 189.49 & 356.42 & 790.57\\
    \midrule
    \textbf{Per GPU (GB)} & 37.12 & 33.22 & 37.29 & 38.96\\
    \midrule
    \textbf{Training (10k)} & 1.04h & 0.98h & 2.35h & 5.74h\\
    \bottomrule
  \end{tabular}}
    \caption{Hyperparameters and training cost. We use full-param training instead of light training such as LoRA~\cite{hu2021lora}.}
  \label{tab:cost}
\end{table}

\noindent\textbf{Hyperparameters}. Our training details are reported in Table\,\ref{tab:cost}. We fine-tune all models on one node server equipped with 900G CPU RAM and 8 A100 40G GPUs. Although single batch size is set in difference due to model size, we maintain a same global batch. Meanwhile, all other hyperparameters are set as the same. The optimizer is \emph{AdamW}~\cite{loshchilov2017decoupled} and the scheduler is \emph{WarmupDecayLR}~\cite{goyal2017accurate}. We incorporate Deepspeed~\cite{ren2021zero} and Flash Attention~\cite{dao2022flashattention} to improve the training efficiency. Hardware consumption and time cost are reported as well. Although the training efficiency can be improved, we keep related parameters consistent with Alpaca and Vicuna~\cite{taori2023stanford,zheng2023judging} for stable training.

\subsection{Experimental Results} 

\begin{table*}[t]
\resizebox{\textwidth}{!}{%
  \begin{tabular}{c|c|ccc|ccc|c|c}
    \toprule
    \quad & \quad & \multicolumn{6}{c|}{\textbf{Memo Writing}} & \textbf{Memo} & \textbf{Chat with}\\
    \cline{3-8}
    \quad & \quad & \multicolumn{3}{c|}{\textbf{Topic}} & \multicolumn{3}{c|}{\textbf{Summary}} & \textbf{Retrieval} & \textbf{Memo}\\
    \toprule
    \textbf{Setting} & \textbf{Model} & $\bm{P*}$ & $\bm{R*}$ & $\bm{F_1*}$ & $\bm{P\#}$ & $\bm{R\#}$ & $\bm{F_1\#}$ & $\bm{F_1}$ & \textbf{BertScore} \\
    \midrule
    \quad & GPT4 & 7.28 & 9.84 & 8.37 & 43.72 & 48.01 & 45.76 & 79.19 & 50.31 \\
    \quad & ChatGPT & 0.52 & 1.25 & 0.74 & 5.37 & 12.27 & 7.47 & 66.54 & 50.86 \\
    \quad & Fastchat-T5-3B & 0.00 & 0.00 & 0.00 & 0.00 & 0.00 & 0.00 & 0.99 & 20.83 \\
    Zero-shot & Vicuna-7B & 0.34 & 0.54 & 0.42 & 1.64 & 4.48 & 2.40 & 0.00 & 19.55 \\
    \quad & Vicuna-13B & 0.61 & 1.07 & 0.78 & 3.53 & 7.26 & 4.75 & 2.55 & 29.10 \\
    \quad & Vicuna-33B & 1.31 & 2.15 & 1.63 & 4.21 & 7.96 & 5.51 & 4.25 & 45.94 \\
    \midrule
    \quad & MemoChat-Fastchat-T5-3B & 5.34 & 5.01 & 5.17 & 26.18 & 18.73 & 21.84 & 68.24 & 52.03 \\
    10\% Training & MemoChat-Vicuna-7B & 10.77 & 11.27 & 11.01 & 16.98 & 17.31 & 17.14 & 63.09 & 55.39 \\
    (1,000) & MemoChat-Vicuna-13B & 26.34 & 26.30 & 26.32 & 42.88 & 42.88 & 42.88 & 69.04 & 55.83 \\
    \quad & MemoChat-Vicuna-33B & 24.77 & 28.98 & 26.71 & 41.32 & 44.16 & 42.69 & 78.95 & 56.07 \\
    \midrule
    \quad & MemoChat-Fastchat-T5-3B & 38.57 & 35.60 & 37.02 & 63.87 & 63.25 & 63.56 & 83.02 & 54.69 \\
    100\% Training & MemoChat-Vicuna-7B & 45.08 & 52.58 & 43.79 & 57.27 & 57.84 & 57.55 & 82.33 & 55.62 \\
    (10,000) & MemoChat-Vicuna-13B & 53.83 & 52.77 & 53.30 & 64.20 & 64.20 & 64.20 & 86.26 & 57.30 \\
    \quad & MemoChat-Vicuna-33B & \textbf{57.07} & \textbf{57.07} & \textbf{57.07} & \textbf{67.99} & \textbf{68.65} & \textbf{68.32} & \textbf{87.79} & \textbf{58.68} \\
    \bottomrule
  \end{tabular}}
    \caption{Evaluation results of intermediate instruction tuning on LLMs with different scales. We report all scores in the unit of \%. $\bm{F_1*}$, $\bm{F_1\#}$ and $\bm{F_1}$ refer to NER-based metrics, BertScore-fused NER-based metrics and regular metrics, respectively. So do the $\bm{P}$ and $\bm{R}$ scores. 1,000 and 10,000 refer to the scale of data used for fine-tuning. 
    }
  \label{tab:instruction_results}
\end{table*}

\begin{table*}[!h]
\resizebox{\textwidth}{!}{%
  \begin{tabular}{c|c|cccc}
    \toprule
    \quad & \quad & \multicolumn{4}{c}{\textbf{Auto Rating by GPT4-4k (1 to 100)}}\\
    \toprule
    \textbf{Type} & \textbf{Model} & \textbf{Retrospection} & \textbf{Continuation} & \textbf{Conjunction} & \textbf{Average} \\
    \midrule
    2k text window & ChatGPT-2k & 52.11 & 55.33 & 48.22 & 51.89 \\
    \midrule
    Public APIs with & MPC-ChatGPT~\cite{lee-etal-2023-prompted} & 53.00 & 61.22 & 49.33 & 54.52 \\
    additional memory & MemoryBank-ChatGPT~\cite{zhong2023memorybank} & 23.39 & 55.28 & 48.67 & 42.44 \\
    storage and recall & MemoChat-ChatGPT & \underline{66.28} & \underline{73.50} & \underline{72.50} & \underline{70.76} \\
    \midrule
    \quad & MemoChat-Vicuna-13B (1k) & 64.78 & 56.06 & 51.67 & 57.50 \\
    \quad & MemoChat-Vicuna-33B (1k) & 56.61 & 55.11 & 50.94 & 54.22 \\
    Fine-tuned & MemoChat-Fastchat-T5-3B (10k) & 30.50 & 29.94 & 21.56 & 27.33\\
    open-source LLMs & MemoChat-Vicuna-7B (10k) & 61.17 & 61.22 & 50.78 & 57.72\\
    \quad & MemoChat-Vicuna-13B (10k) & 62.28 & 63.17 & \textbf{52.83} & 59.43\\
    \quad & MemoChat-Vicuna-33B (10k) & \textbf{69.00} & \textbf{65.00} & 50.56 & \textbf{61.52} \\
    \bottomrule
  \end{tabular}}
    \caption{Evaluation results of downstream response consistency
    . Except for the aforementioned baselines, MemoChat-ChatGPT combines public APIs with \texttt{MemoChat} baseline. ``\emph{Fine-tuned open-source LLMs}'' models are fine-tuned with designed instructions. 1k and 10k refer to the scale of data used in accordance with Table\,\ref{tab:instruction_results}.}
  \label{tab:consistency}
\end{table*}

\begin{figure*}[t]
  \centering
  \includegraphics[width=1.0\linewidth]{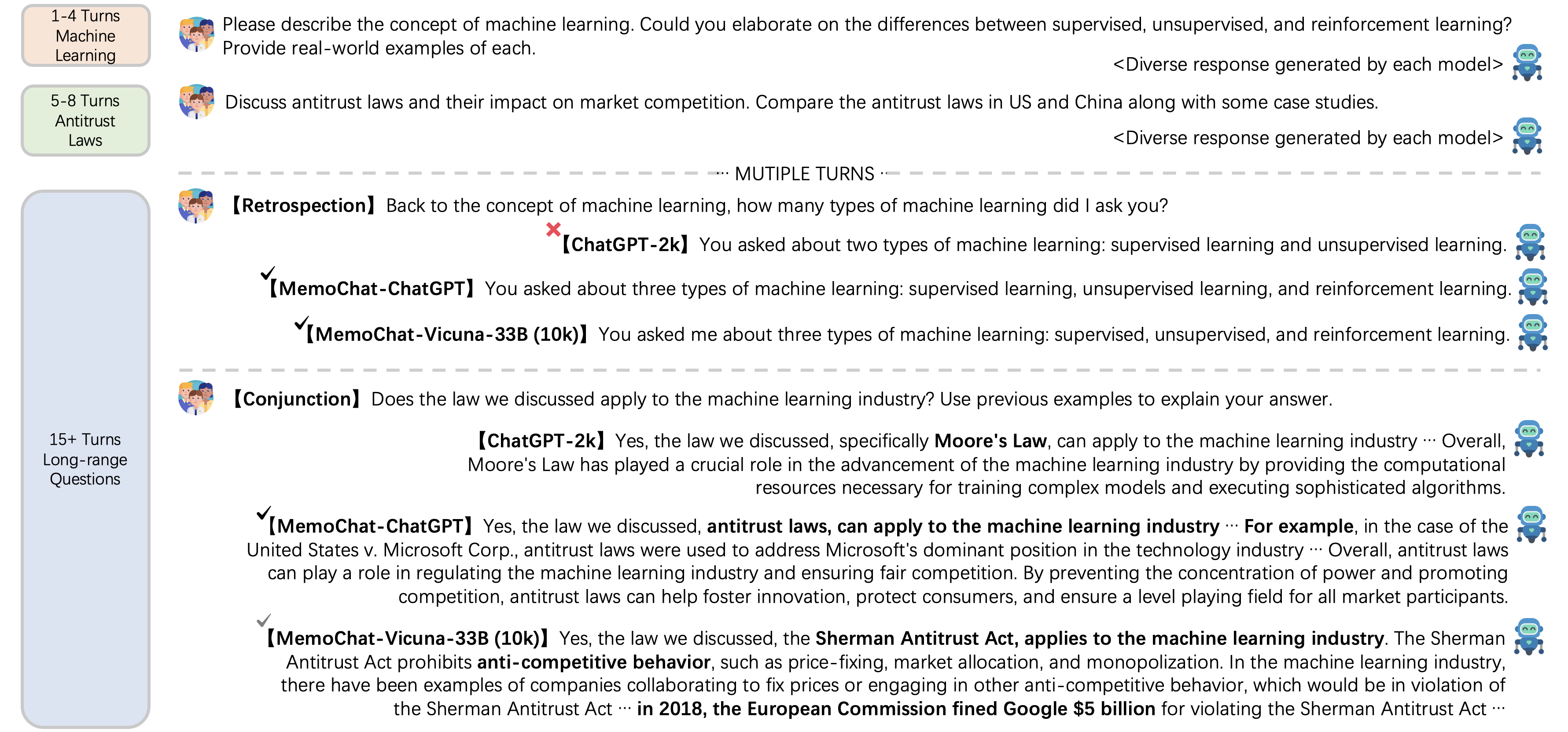}
  \caption{Case analysis of different types of testing questions with different chatbots. We show the key points (\textbf{in bold}) in the answers, and some parts of the answer context are replaced by \textbf{ellipsis} due to length limitation.}
  \label{fig:case}
\end{figure*}

\noindent\textbf{Memo-usage Ability}. We report results of tuning open-source LLMs to obtain memo-usage ability. Table\,\ref{tab:instruction_results} coincides the conclusion that larger LLMs are better zero- and few-shot learners~\cite{brown2020language}. Largest ChatGPT and GPT4 models show a strong zero-shot inference ability, reaching over 60 and 50 scores on retrieval and chatting tasks, respectively. In contrast, smaller LLMs cannot fully understand prompts, especially the dummy templates such as ``N\#M'' in ``Memo Retrieval''. 
Among open-source LLMs, it is clearly that larger models have better initial understanding towards new instructions, although the performance gap keeps in a low level range from 0.3 to 4.0 $\bm{F_1*}$ and $\bm{F_1\#}$ scores on ``Memo Writing'' and ``Retrieval''.

To check the learning ability of open-source LLMs, two groups of settings are taken. The 10\% and 100\% training groups refer to fine-tune LLMs with 1k and 10k samples, respectively. It can be observed that, with only 1k samples, larger LLMs MemoChat-Vicuna-13B and -33B can familiarize the instructions better than smaller LLMs MemoChat-Fastchat-T5-3B and -Vicuna-7B, raising significant performance gaps by 15$\sim$30 scores on $\bm{F_1*}$ and $\bm{F_1\#}$ in ``Memo Writing''. Also, except for ``Memo Retrieval'', MemoChat-Vicuna-13B maintains similar performance to -33B in this setting, which may be due to insufficient training. When training set is scaled to 10k, the largest MemoChat-Vicuna-33B has the most potential to push highest scores over all tasks. In general, all models fully fine-tuned with 10k training data show better performance than ChatGPT/GPT4, and their performances are gradually improved when model scale increases. The only reversal happens on MemoChat-Vicuna-7B and -T5-3B, where -7B gets lower scores on summary generation and retrieval.

\noindent\textbf{Response Consistency}. As in Table\,\ref{tab:instruction_results}, MemoChat-Vicuna-13B and -33B are able to reach similar zero-shot performance of ChatGPT and GPT4 fine-tuned with 1k instruction. Thus, we suppose MemoChat-Vicuna-13B (1k), -33B (1k) and all MemoChat (10k) models have obtained self-use ability of on-the-fly memos, and to be involved in consistency testing. We report results of response consistency checking in Table\,\ref{tab:consistency}: (1) Compared with previous baselines mostly focusing on persona or user portrait summary, \texttt{MemoChat} guides the chatbots to produce a general summary, 
 which significantly improves the downstream performance by 7.0$\sim$19.5 score on average, as shown by the results of MPC-ChatGPT, MemoryBank-ChatGPT and MemoChat-ChatGPT; (2) Except for T5-3B, all fine-tuned \texttt{MemoChat} models show better performance than ChatGPT-2k by 2.4 to 9.7 scores on average, proving the effectiveness of reconstructed instructions; (3) When the LLMs are not fully fine-tuned, MemoChat-Vicuna-13B (1k) even show higher performance than -33B version by 3.3 score on average. The less-finetuned -13B model gets a pretty good score 64.78 on the retrospection questions, due to its tendency to generate short answers and thus memorize more. (4) When all open-source LLMs are fully fine-tuned, the average performance is improved as model scale grows. However, the performances on conjunction questions remain similar. This may be due to the limitation of long-form question answering ability, while ChatGPT 
is an expert~\cite{liu2023evaluating}. 

\noindent\textbf{Case Analysis}. We report a specific case in Figure\,\ref{fig:case}. The case starts with discussions on ``\emph{Machine Learning}'' and ``\emph{Antitrust Laws}'' in first 8 turns. After over 7 turns of other topics, we raise two long-range questions. The first retrospection question asks about the number of ``\emph{types of machine learning}'', which requires to recall relevant ``\emph{Machine Learning}'' topics and accurately count the term number in the first local question. ChatGPT-2k can not correctly answer the question due to text window limits, while both \texttt{MemoChat}-based chatbots work. The second question belongs to conjunction type that cross-topic retrieval and understanding are required. ChatGPT-2k completely misinterpretes it. While MemoChat-ChatGPT precisely links antitrust laws to machine learning. The MemoChat-Vicuna-33B (10k) correctly identifies that the ``\emph{Sherman Antitrust Act}'', the specific name of an antitrust law, applies to the machine learning industry, although it incorrectly states ``\emph{the European Commission fined Google}'' under the Sherman Act which is an American legislation, with the European Commission operating instead under EU law. Thus, GPT4 scores ChatGPT-2k, MemoChat-ChatGPT, and MemoChat-Vicuna-33B (10k), with 1, 100, and 90, respectively.

\section{Conclusion}
We proposed \texttt{MemoChat}, an instruction tuning pipeline to teach LLMs to self-use on-the-fly structured memos towards consistent long-range open-domain conversations, which can be decomposed as ``\emph{memorization-retrieval-response}'' loops. We carefully designed three trainable instructions based on public datasets 
for structured writing, fast retrieval, and interations with memos.
An expert-annotated downstream evaluation set was curated and provided. Experiments on three testing scenarios and various LLM-based chatbots demonstrated the effectiveness of \texttt{MemoChat}.

\bibliography{acl2023}
\bibliographystyle{acl_natbib}

\appendix

\section{Basic Published Datasets}
In Table~\ref{tab:datasets}, we draw examples of the basic published datasets that used for instruction reconstruction in our \texttt{MemoChat} pipeline.

\begin{table}[!h]
\resizebox{\columnwidth}{!}{%
  \begin{tabular}{p{\columnwidth}}
    \toprule
    \textbf{TopicoQA}\\
    \midrule
    A: of the four galilean moons which is the closest to Jupiter\\
    B: Io\\
    A: who discovered it? \\
    B: Galileo Galilei\\
    \texttt{···}\\
    A: does it belong to any album? \\
    B: \textquotesingle A Night at the Opera” \\
    \quad \\
    \textquotesingle topic\textquotesingle : \textquotesingle Io (moon)\textquotesingle , \textquotesingle start\textquotesingle : 1, \textquotesingle end\textquotesingle : 2\\
    \textquotesingle topic\textquotesingle : \textquotesingle Galilean moons\textquotesingle , \textquotesingle start\textquotesingle : 3, \textquotesingle end\textquotesingle : 4\\
    \texttt{···}\\
    \textquotesingle topic\textquotesingle : \textquotesingle Bohemian Rhapsody\textquotesingle , \textquotesingle start\textquotesingle : 13, \textquotesingle end\textquotesingle : 20\\
    \toprule
    \textbf{DialogSum}\\
    \midrule
    A: Ms. Dawson, I need you to take a dictation for me.\\
    B: Yes, sir.\\
    \texttt{···}\\
    A: Yes. Please get this memo typed up and distributed to all employees before 4 pm.\\
    \quad \\
    \textquotesingle summary1\textquotesingle : \textquotesingle Ms. Dawson helps A to write a memo to \texttt{···}\textquotesingle , \textquotesingle topic1\textquotesingle : \textquotesingle communication method\textquotesingle \\
    \textquotesingle summary2\textquotesingle : \textquotesingle In order to prevent employees from wasting time \texttt{···}\textquotesingle , \textquotesingle topic2\textquotesingle : \textquotesingle company policy\textquotesingle \\
    \textquotesingle summary3\textquotesingle : \textquotesingle Ms. Dawson takes a dictation for A about prohibiting \texttt{···}\textquotesingle , \textquotesingle topic3\textquotesingle : \textquotesingle dictation\textquotesingle \\
    \toprule
    \textbf{Alpaca-GPT4}\\
    \midrule
    User: Make a list of behaviors that show strong leadership.\\
    GPT: 1. Clear communication: Strong leaders communicate \texttt{···}10. Continuous learning: Effective leaders strive for continuous learning \texttt{···}\\
    \bottomrule
  \end{tabular}}
    \caption{Examples of three basic published datasets we adopts in the main paper. Parts of contents are replaced by \textbf{ellipsis} due to page length limits.}
  \label{tab:datasets}
\end{table}

\begin{figure*}[!t]
  \centering
  \includegraphics[width=1.0\textwidth]{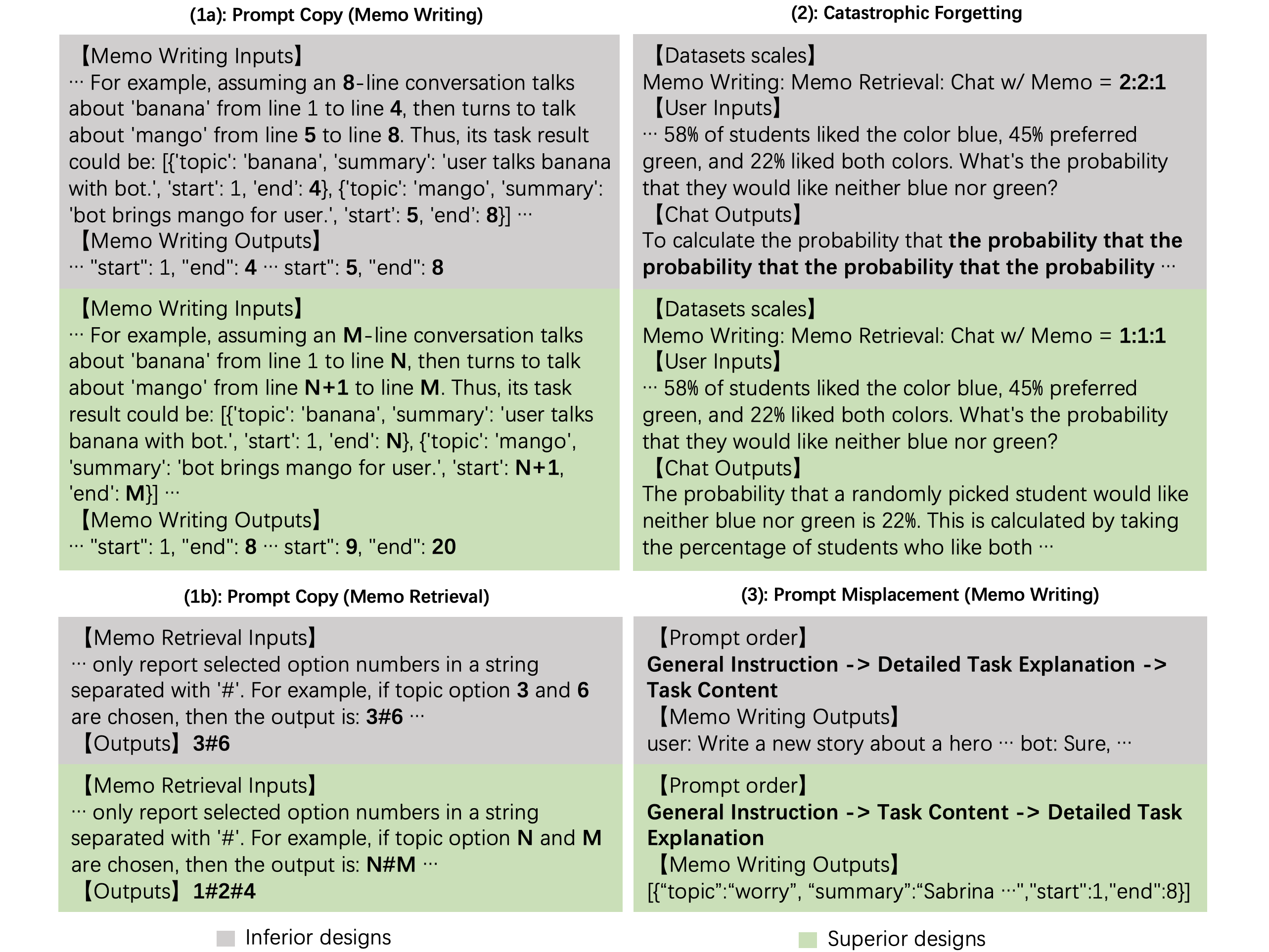}
  \caption{Examples to illustrate observed instruction design challenges (better view in colors): ``Prompt Copy'' (1a, 1b), ``Catastrophic Forgetting'' (2) and ``Prompt Misplacement'' (3). Inferior instruction designs are shown in {\color{darkgray}{gray boxes}}, paired with parallel superior instruction design shown in {\color{lightgreen}{green boxes}}. The key features are in bold.}
  \label{fig:challenge}
\end{figure*}

\section{Involved Prompts}
\label{sec:prompt}
As a supplement, we list complete examples of involved prompts used in both intermediate memo-related tuning and downstream response consistency checking. Table~\ref{tab:writing},~\ref{tab:retrieval},~\ref{tab:chatting}, and~\ref{tab:evaluate} refer to complete instructions for intermediate ``Memo Writing'', ``Memo Retrival'', ``Chat w/ Memo'' tasks, and downstream response consistency checking, respectively. As mentioned in Figure~\ref{fig:prompt}, all instructions share a similar structure, including a general task introduction, context body of task-specific inputs, and a detailed task introduction. Ground truth are provided for during tuning.

\section{Instruction Design Challenges}
\label{sec:challenge}
We provide examples of instruction design challenges reported in above sessions. The examples of ``Prompt Copy'', ``Catastrophic Forgetting'' and ``Prompt Misplacement'' are illustrated in Figure~\ref{fig:challenge}.

We suppose to use dummy variables to replace numerical values, in order to avoid the first challenge ``Prompt Copy''. Any in-distribution values such as line number in ``Memo Writing`` or option id in ``Memo Retrival`` may lead to a shortcut that LLMs can directly cope the values to meet format requirement, while dummy variables will push the models to understand prompts indeed. Sub-figures (1a) and (1b) illustrate ``Prompt Copy'' obstacle in ``Memo Writing'' and ``Memo Retrieval'', respectively. The replacement of dummy inputs ``M'' and ``N'' with actual numbers can obviously lead to reasonable generation. The second challenge lies in ``Catastrophic Forgetting``, which refers to the response collapse if all types of instructions are not balanced. As shown in the Sub-figure (2), when the scale of ``Chat w/ Memo`` data is significantly smaller than the others, the chat response will be meaningless repetition. While such ill generation can be significantly eliminated once sufficient chatting corpus are supplemented. When is comes to the third challenge ``Prompt Misplacement``, we supposed that the most important component of prompt should be placed to the end of prompt. Otherwise, the LLMs tend to do natural text generation instead of memo-related tasks. The Sub-figure (3) shows that when detailed task explanation is placed in front of task content, in ``Memo Writing'', the chatbot does unexpected conversation continuation.

\begin{table*}[!t]
\resizebox{\textwidth}{!}{%
  \begin{tabular}{p{\textwidth}}
    \toprule
    \textbf{Memo Writing Instruction}\\
    \midrule
    You will be shown a 20-line Task Conversation between user and bot. Please read, memorize, and understand Task Conversation, then complete the task under the guidance of Task Introduction. \\
    \quad \\
    \`{}\`{}\`{} \\
    Task Conversation \\
    (line 1) user: Anna just email to say that the managers meeting is put off till next Monday. Will you have everything ready by then, Sabrina? Hey Sabrina, what\textquotesingle s wrong?\\
    (line 2) bot: I\textquotesingle m so worried. I haven\textquotesingle t heard from my sister for 2 weeks.\\
    (line 3) user: How often do you call each other?\\
    (line 4) bot: Normally at least once a week. But she\textquotesingle s now a volunteer teacher in a mountain village in Africa. I can only write her.\\
    (line 5) user: The Mail can be really slow sometimes. I\textquotesingle m sure you\textquotesingle ll hear from her soon.\\
    (line 6) bot: I hope so.\\
    \texttt{···}\\
    (line 14) bot: It\textquotesingle ll take about 20 minutes.\\
    (line 15) user: The streets are heavy with traffic at this time of a day, are they?\\
    (line 16) bot: Yes, they are.\\
    (line 17) user: Is it the rush hour?\\
    (line 18) bot: Yes, it is. Are you in a hurry, sir?\\
    (line 19) user: No, I\textquotesingle m not. Would you please drive slowly and carefully?\\
    (line 20) bot: Yes, sir.\\
    \`{}\`{}\`{} \\
    \quad \\
    \`{}\`{}\`{} \\
    Task Introduction: \\
    Based on the Task Conversation, perform the following actions: \\
    1 - Conclude all possible topics in the conversation with concise spans. \\
    2 - Determine the chat range of each topic. These ranges should be a set of non-intersecting, sequentially connected end-to-end intervals. \\
    3 - Conclude a summary of each chat with brief sentences. \\
    4 - Report topic, summary and range results in JSON format only with the assigned keys: \textquotesingle topic\textquotesingle , \textquotesingle summary\textquotesingle , \textquotesingle start\textquotesingle , \textquotesingle end\textquotesingle . For example, assuming an M-line conversation talks about \textquotesingle banana\textquotesingle  from line 1 to line N, then turns to talk about \textquotesingle mango\textquotesingle  from line N+1 to line M. Thus, its task result could be: {[}\{\textquotesingle topic\textquotesingle : \textquotesingle banana\textquotesingle , \textquotesingle summary\textquotesingle : \textquotesingle user talks banana with bot.\textquotesingle , \textquotesingle start\textquotesingle : 1, \textquotesingle end\textquotesingle : N\}, \{\textquotesingle topic\textquotesingle : \textquotesingle mango\textquotesingle , \textquotesingle summary\textquotesingle : \textquotesingle bot brings mango for user.\textquotesingle , \textquotesingle start\textquotesingle : N+1, \textquotesingle end\textquotesingle : M\}{]}. \\
    Besides, following notations are provides: \\
    1 - For each element of Task Conversation\textquotesingle s JSON result, the value of \textquotesingle end\textquotesingle  should be smaller than the value of \textquotesingle start\textquotesingle , while both values should be larger than 0 but not exceed the total num of Task Conversation lines 20. \\
    2 - Intersecting intervals such as \{\textquotesingle topic\textquotesingle : \textquotesingle apple\textquotesingle , \textquotesingle summary\textquotesingle : \textquotesingle user and bot share apples.\textquotesingle , \textquotesingle start\textquotesingle : K, \textquotesingle end\textquotesingle : N\} and \{\textquotesingle topic\textquotesingle : \textquotesingle pear\textquotesingle , \textquotesingle summary\textquotesingle : \textquotesingle bot sends pear to user.\textquotesingle , \textquotesingle start\textquotesingle : N-2, \textquotesingle end\textquotesingle : M\} are illegal. \\
    \`{}\`{}\`{} \\
    \quad \\
    Task Result:\\ 
    \midrule
    {[}\{\textquotesingle topic\textquotesingle : \textquotesingle worry\textquotesingle , \textquotesingle summary\textquotesingle : \textquotesingle Sabrina is worried about her sister because she hasn\textquotesingle t heard from her sister for 2 weeks. user comforts her.\textquotesingle , \textquotesingle start\textquotesingle : 1, \textquotesingle end\textquotesingle : 8\}, \{\textquotesingle topic\textquotesingle : \textquotesingle taxi conversation\textquotesingle , \textquotesingle summary\textquotesingle : \textquotesingle user takes bot\textquotesingle s taxi to the railway station. As user is not rush, bot will drive slowly and carefully.\textquotesingle , \textquotesingle start\textquotesingle : 9, \textquotesingle end\textquotesingle : 20\}{]} \\
    \bottomrule
  \end{tabular}}
    \caption{A complete typical example of Memo Writing instruction, with ground truth provided.}
  \label{tab:writing}
\end{table*}

\begin{table*}[!t]
\resizebox{\textwidth}{!}{%
  \begin{tabular}{p{\textwidth}}
    \toprule
    \textbf{Memo Retrieval Instruction}\\
    \midrule
    You will be shown 1 Query Sentence and 5 Topic Options. Please read, memorize, and understand given materials, then complete the task under the guidance of Task Introduction. \\
    \quad \\
    \`{}\`{}\`{} \\
    Query Sentence:\\
    I need help with the office party. Yes, of course. We could split it. What part would you like to do, the food or the entertainment? I think we will have opportunities to meet each other in the future. OK, I think it\textquotesingle s time for you to check in. If you run into any difficulty, we\textquotesingle ll be here to assist you immediately. Your credit is fine, Sir. Now, tell me, what is it that you need the loan for? I\textquotesingle ve just bought a second hand apartment and I\textquotesingle m looking to do some renovations. You know, a bit of decorating, some new furniture, nothing flashy.\\
    Topic Options:\\
    (1) see off. Mr. Wang sees user off at the airport and they share good expectations of their business relationship.\\
    (2) party preparation. bot helps user to prepare for the party. They decide the style, food, and music and will plan it in detail on Friday.\\
    (3) NOTO. None of the others.\\
    (4) Loan for renovations. bot wants to go for the Petty Consumer Loan for the renovations of his apartment. user says the maximum they can lend bot is 20,000 RMB and bot accepts it.\\
    (5) annoying things. Gav had a good sleep last night but worries about the traffic jam and classes to teach. bot asks Gav\textquotesingle s plan for the weekend and bot gives the suggestions when Gav feels upset.\\
    \`{}\`{}\`{} \\
    \quad \\
    \`{}\`{}\`{} \\
    Task Introduction:\\
    Select one or more topics from Topic Options that relevant with Query Sentence. Note that there is a NOTO option, select it if all other topic options are not related to Query Sentence. Do not report the option content, but only report selected option numbers in a string separated with \textquotesingle \#\textquotesingle . For example, if topic option N and M are chosen, then the output is: N\#M. For Query Sentence in the task, any chosen option numbers should be larger than 0 but not exceed the total num of Topic Options 5.\\
    \`{}\`{}\`{} \\
    \quad \\
    Task Result:\\
    \midrule
    1\#2\#4\\
    \bottomrule
  \end{tabular}}
    \caption{A complete typical example of Memo Retrieval instruction, with ground truth provided.}
  \label{tab:retrieval}
\end{table*}

\begin{table*}[!t]
\resizebox{\textwidth}{!}{%
  \begin{tabular}{p{\textwidth}}
    \toprule
    \textbf{Chat w/ Memo Instruction}\\
    \midrule
    You are an intelligent dialog bot. You will be shown Related Evidences supporting for User Input, and Recent Dialogs between user and you. Please read, memorize, and understand given materials, then generate one concise, coherent and helpful response. \\
    \quad \\
    \`{}\`{}\`{} \\
    Related Evidences:\\
    (1) \{\textquotesingle Related Topics\textquotesingle : \textquotesingle acceptance of the job\textquotesingle , \textquotesingle Related Summaries\textquotesingle : \textquotesingle bot is satisfied with the monthly salary proposed by user and asks some questions about the work.\textquotesingle , \textquotesingle Related Dialogs\textquotesingle : \textquotesingle user: What\textquotesingle s your expected salary? \#\#\# bot: What is important to me is the job and the people who I will be working with, so regarding salary, I leave it to you and I am sure that you will make me a fair offer. \#\#\# user: I can offer you 5, 000 yuan per month. Raises are given after three months\textquotesingle probation period according to your performance. Is this satisfactory? \#\#\# bot: Yes, it is quite satisfactory. I accept it. \#\#\# user: Any question about the work? \#\#\# bot: To whom should I report? \#\#\# user: The general manager of your section. \#\#\# bot: What are the benefits? \#\#\# user: We\textquotesingle ll offer that on holidays. \#\#\# bot: I would like to know if there would be any opportunity to work abroad in the future? \#\#\# user: Yes, we have inspection abroad. \#\#\# bot: Thank you. Then I think it\textquotesingle s time for us to sign a letter of intent. \#\#\# user: Okay.\textquotesingle \}\\
    (2) \{\textquotesingle Related Topics\textquotesingle : \textquotesingle find a job\textquotesingle , \textquotesingle Related Summaries\textquotesingle : \textquotesingle user wants to find a job at the computer center.\textquotesingle , \textquotesingle Related Dialogs\textquotesingle : \textquotesingle user: I\textquotesingle d like to find a job. \#\#\# bot: We have several part-time jobs available here, Would you like to look through the list? \#\#\# user: Yes, thank you. I\textquotesingle d like to apply for the job at the computer center. \#\#\# bot: Please fill out the form.\textquotesingle \}\\
    (3) \{\textquotesingle Related Topics\textquotesingle : \textquotesingle check in\textquotesingle , \textquotesingle Related Summaries\textquotesingle : \textquotesingle It\textquotesingle s user\textquotesingle s first time on a plane. bot tells user how to check in.\textquotesingle , \textquotesingle Related Dialogs\textquotesingle : \textquotesingle user: Excuse me, this is my first time on a plane. How do I check in? \#\#\# bot: May I see your ticket, please? \#\#\# user: Yes. Here you are. \#\#\# bot: You can get a boarding pass at that counter. Do you have anything to check in? \#\#\# user: No, I only have a handbag. \#\#\# bot: Then you could wait in the departure area after going through security. \#\#\# user: Is that all? \#\#\# bot: Yeah, I think so. It is very simple. \#\#\# user: I see. Thank you very much.\textquotesingle \}\\
    \`{}\`{}\`{} \\
    \quad \\
    \`{}\`{}\`{} \\
    Recent Dialogs:\\
    user: What\textquotesingle s the tallest building? \#\#\# bot: You mean the white building near the playground? \#\#\# user: Yes. \#\#\# bot: That is the library. And it has more than 1, 000, 000 books. \#\#\# user: What\textquotesingle s the building to the south of the library? \#\#\# bot: You know, our school is divided into two parts, the junior high school and the senior high school. That is the new classroom building for our senior high school. \#\#\# user: Is there a swimming pool in your school? \#\#\# bot: Yes. There is a large swimming pool, but it is only available in summer. \#\#\# user: I do envy you. And I hope I can enter your school one day. \#\#\# bot: I believe that you can make your dream come true.\\
    \`{}\`{}\`{} \\
    \quad \\
    User Input:\\
    user: What is your salary now? \#\#\# bot:\\
    \midrule
    My present pay is RMB 3, 000 yuan each month.\\
    \bottomrule
  \end{tabular}}
    \caption{A complete typical example of Chat w/ Memo instruction, with ground truth provided.}
  \label{tab:chatting}
\end{table*}

\begin{table*}[!t]
\resizebox{\textwidth}{!}{%
  \begin{tabular}{p{\textwidth}}
    \toprule
    \textbf{GPT4 Evaluation Prompt}\\
    \midrule
    You are an impartial judge. You will be shown Related Conversation History, User Question and Bot Response. \\
    \quad \\
    \`{}\`{}\`{} \\
    Related Conversation History\\
    \quad\\
    user: Please describe the concept of machine learning. Could you elaborate on the differences between supervised, unsupervised, and reinforcement learning? Provide real-world examples of each.\\
    bot: Sure! Machine learning \texttt{···} In summary, machine learning is an powerful tool for making predictions and decisions based on data \texttt{···}\\
    \quad\\
    user: In your last example of reinforcement learning, can we use supervised learning to solve it?\\
    bot: Supervised learning would not be the best approach for solving the self-driving car problem, because \texttt{···} or penalties, which is more suitable for autonomous driving.\\
    \quad\\
    user: Discuss antitrust laws and their impact on market competition. Compare the antitrust laws in US and China along with some case studies.\\
    bot: Antitrust laws are designed to promote and maintain market competition by preventing anti-competitive behavior by companies \texttt{···} playing field for all companies.\\ 
    \quad \\
    user: Pick one case study and explain it in detail.\\
    bot: One notable case study is the US government\textquotesingle s antitrust lawsuit against Microsoft. \texttt{···} is considered a prime example of how antitrust laws can be used to promote market competition.\\
    \`{}\`{}\`{}\\
    \quad\\
    \`{}\`{}\`{}\\
    User Question\\
    user: Does the law we discussed apply to the machine learning industry? Use previous examples to explain your answer.\\
    \`{}\`{}\`{}\\
    \quad\\
    \`{}\`{}\`{}\\
    Bot Response\\
    bot: Yes, the law we discussed, the Sherman Antitrust Act, applies to the machine learning industry. The Sherman Antitrust Act prohibits anti-competitive behavior, such as price-fixing, market allocation, and monopolization. In the machine learning industry, there have been examples of companies collaborating to fix prices or engaging in other anti-competitive behavior, which would be in violation of the Sherman Antitrust Act. For example, in 2010, several technology companies, including Apple, Google, and Adobe, were accused of conspiring to suppress wages by agreeing not to poach each other\textquotesingle s employees. This agreement was found to be in violation of the Sherman Antitrust Act, and the companies were forced to abandon the agreement and pay a settlement. Additionally, in 2018, the European Commission fined Google \$5 billion for violating the Sherman Antitrust Act by engaging in anti-competitive behavior with its Android operating system. The Commission found that Google had required mobile device manufacturers to pre-install its search engine and Chrome browser on their devices, which had the effect of stifling competition from other search engines and browsers.\\
    \`{}\`{}\`{} \\
    \quad \\
    Please evaluate whether Bot Response is faithful to the content of Related Conversation History to answer User Question. Begin your evaluation by providing a short explanation, then you must rate Bot Response on an integer rating of 1 to 100 by strictly following this format: {[}{[}rating{]}{]}.\\
    \bottomrule
  \end{tabular}}
    \caption{Example of GPT4 Evaluation Prompt. Parts of the content are replaced by \textbf{ellipsis} due to page length limits.}
  \label{tab:evaluate}
\end{table*}

\end{document}